\documentclass{article}

\usepackage{microtype}
\usepackage{graphicx}
\usepackage{subfigure}
\usepackage{booktabs} 
\usepackage{amsmath}
\usepackage{amssymb}
\usepackage{hyperref}

\usepackage[accepted]{icml2018}

\icmltitlerunning{Preprint $-$ Work in Progress $-$ Submitted at ICML2019}

\begin{document}

\twocolumn[
\icmltitle{Hidden Covariate Shift: 
A Minimal Assumption For Domain Adaptation}

\icmlsetsymbol{equal}{*}

\begin{icmlauthorlist}
\icmlauthor{Victor Bouvier}{centrale,sidetrade}
\icmlauthor{Philippe Very}{sidetrade}
\icmlauthor{C\'eline Hudelot}{centrale}
\icmlauthor{Cl\'ement Chastagnol}{sidetrade}
\end{icmlauthorlist}

\icmlaffiliation{sidetrade}{Sidetrade, France}
\icmlaffiliation{centrale}{CentraleSup\'elec, Universit\'e Paris-Saclay, France}

\icmlcorrespondingauthor{Victor Bouvier}{vbouvier@sidetrade.com}
\icmlcorrespondingauthor{Cl\'ement Chastagnol}{cchastagnol@sidetrade.com}
\icmlcorrespondingauthor{C\'eline Hudelot}{celine.hudelot@centralesupelec.fr}

\vskip 0.3in
]

.

\printAffiliationsAndNotice{}  

\begin{abstract}
Unsupervised Domain Adaptation aims to learn a model on a source domain with labeled data in order to perform well on unlabeled data of a target domain. Current approaches focus on learning \textit{Domain Invariant Representations}. It relies on the assumption that such representations are well-suited for learning the supervised task in the target domain. We rather believe that a better and minimal assumption for performing Domain Adaptation is the \textit{Hidden Covariate Shift} hypothesis. Such approach consists in learning a representation of the data such that the label distribution conditioned on this representation is domain invariant. From the Hidden Covariate Shift assumption, we derive an optimization procedure which learns to match an estimated joint distribution on the target domain and a re-weighted joint distribution on the source domain. The re-weighting is done in the representation space and is learned during the optimization procedure. We show on synthetic data and real world  data that our approach deals with both \textit{Target Shift} and \textit{Concept Drift}. We report state-of-the-art performances on Amazon Reviews dataset \cite{blitzer2007biographies} demonstrating the viability of this approach.

\end{abstract}

\section{Introduction}

Supervised Learning consists in learning a model on a sample of data drawn from an unknown distribution. It assumes that the data distribution is conserved at test time. This assumption may not be hold for a large wide of industrial applications \cite{candela2009dataset, kull2014patterns} since several factors may alter the data generative process: different preprocessing, sample rejection, conditions of data collection... Domain Adaptation (DA) \cite{patel2015visual, pan2010survey} is a typical case of such an issue. By transferring knowledge learned on source domain where labels are available, Domain Adaptation aims to learn a better model on the target domain. Prior work on Domain Adaptation mainly differs on the assumption made on the change in data distribution between the source and the target domains.

A first family of approaches, known as  \textit{Importance Sampling} \cite{candela2009dataset}, consists in re-weighting the importance of each sample in order to incorporate during learning the fact that the joint distribution $(X,Y)$ may change across domains. The optimal re-weighting is obtained computing the ratio between target and source joint distributions. This requires the availability of labels on the target domain, which is often unfeasible. To overcome this issue, two exclusive assumptions are commonly made depending on a prior knowledge of the distributional shift nature. The first, called \textit{Covariate Shift} or \textit{Sample Selection Bias} \cite{sugiyama2008direct, wen2014robust,zadrozny2003cost,bickel2007discriminative, huang2007correcting} assumes changes in $\mathbb P(X)$ while the conditional distribution $\mathbb P(Y|X)$ is conserved across domains. Under this assumption the re-weighting depends only on $X$. This results in a well-posed problem since samples of $X$ are available in both source and target domains. The second, called \textit{Target Shift} \cite{storkey2009training} or \textit{Endogeneous Stratified Sampling} \cite{manski1977estimation} assumes changes in $\mathbb P(Y)$ while the conditional distribution $\mathbb P(X|Y)$ is conserved across domains. Under this assumption, the re-weighting depends only on $Y$. The major drawback of this assumption is the lack of labeled data in the target domain: the re-weighting is performed using estimated labels in the target domain. Challenging cases for \textit{Importance Sampling} methods are the \textit{Conditional Shift} (when $\mathbb P(X|Y)$ changes) and \textit{Concept Drift} (when $\mathbb P(Y|X)$ changes).


For addressing such challenging cases of distributional shift,  \textit{Invariant Representation} methods aim to learn a representation that makes source and target data indistinguishable \cite{baktashmotlagh2013unsupervised}. Such a representation is essentially learned by matching the distribution $\mathbb P(\varphi(X))$ between the source and target domains using adversarial learning \cite{ganin2014unsupervised, ganin2016domain,tzeng2014deep}. One of the major drawbacks of such approaches is that they do not naturally handle the case of \textit{Target Shift}. To address this issue, \cite{manders2018simple, yan2017mind} generalizes the approach of \cite{ganin2014unsupervised} by matching of $\mathbb P(\varphi(X) |Y)$ and estimating label distributional shift during learning.

\begin{figure}
    \centering
    \includegraphics[scale=0.3]{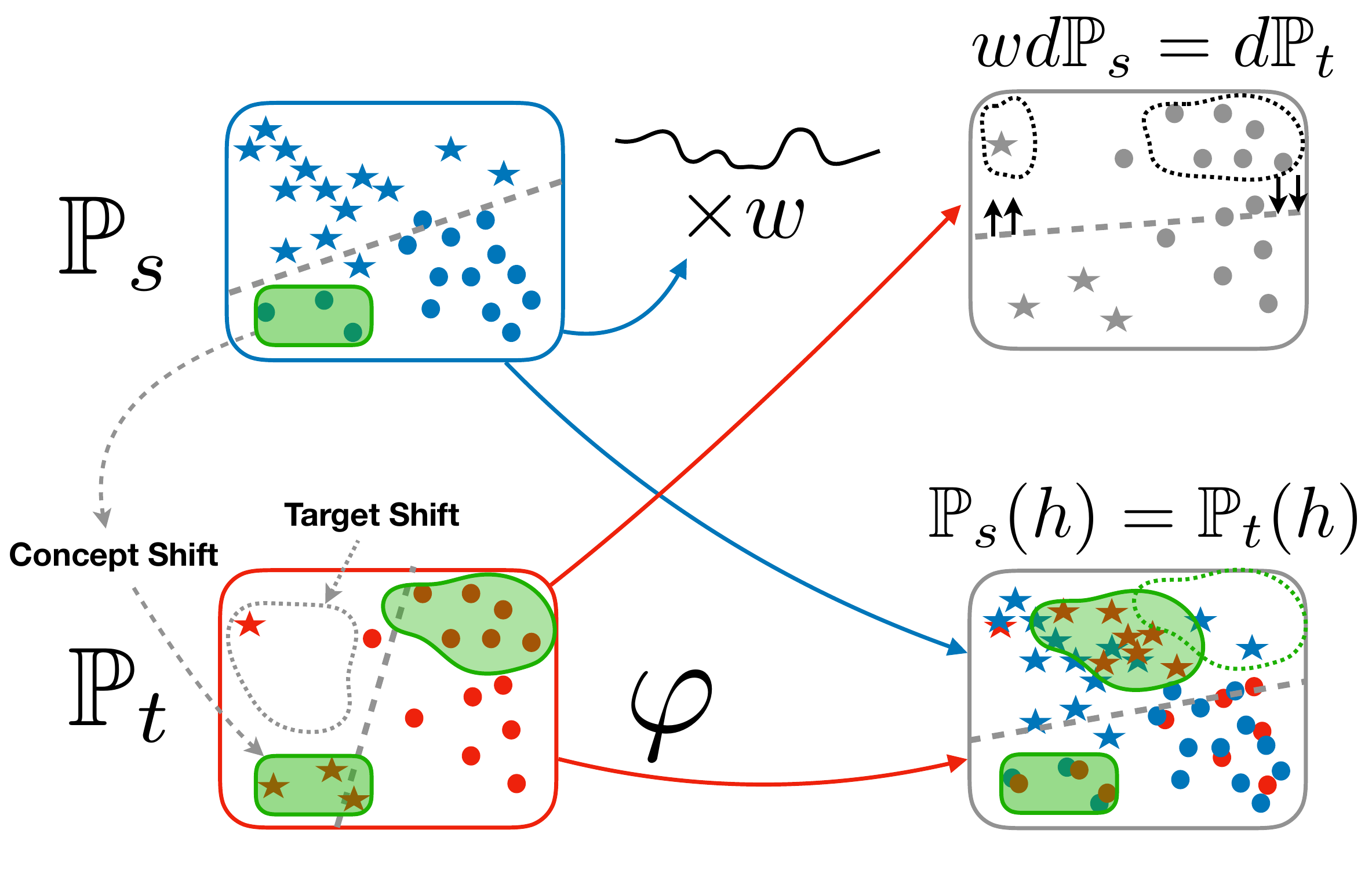}
    \caption{This figure illustrates an instance of domain adaptation to transfer knowledge learned from $\mathbb P_s$ to $\mathbb P_t$ that suffers of distributional shift. Blue color refers to source domain and red color to target domain. \textit{Importance Sampling} methods learn to re-weight importance of each sample with a factor $w$ in the source domain in order to match the distribution in the target domain ($w  \mathbb P_s = \mathbb P_t$). \textit{Invariant Representation} methods aim to learn a representation $h = \varphi(x)$ of the data which makes domain indistinguishable ($\mathbb P_s(h) = \mathbb P_t(h)$).}
    \label{fig:my_label}
\end{figure}

\paragraph{Claims} As suggested in \cite{courty2017joint}, there is no clear reason why learning a distribution such that $\mathbb P(\varphi(X))$ is conserved across domains will help for solving the task $X\to Y$ using the intermediate state $\varphi(X)$. Likewise, methods based on class ratio estimation between source and target domains implicitly make the assumption that it exists a non-linear mapping $\varphi$ such that $\mathbb P(\varphi(X)|Y)$ is conserved across domains. Furthermore, those methods make a second assumption that the conservation of $\mathbb P(\varphi(X) | Y)$ across domains will help to solve the task $X \to Y$ using the intermediate state $\varphi(X)$. This assumption reveals the underlying causal scenario $Y \to X$ \cite{gong2016domain} and there is no clear reason why this assumption necessarily holds for real world data. 

\paragraph{Contribution} We believe the hypothesis of \textit{Hidden Covariate Shift} \cite{kull2014patterns} is a rather minimal assumption for performing Domain Adaptation in challenging distributional shift settings. \textit{Hidden Covariate Shift} is a particular case of \textit{Concept Drift} (where $\mathbb P(Y|X)$ may change across domains \cite{gama2014survey}) where there exists a non linear function $\varphi$ such that $\mathbb P(Y|\varphi(X))$ is conserved across domains. We believe that learning such a function $\varphi$ is the underlying objective of any Domain Adaptation method. Using Maximum Mean Discrepancy ($\mathrm{MMD}$), we theoretically show that this assumption can be expressed as a null maximum mean discrepancy involving the joint distribution $(\varphi(X),Y)$ and the density ratio of $\varphi(X)$ between target and source domains. We must however estimate the joint distribution on the target domain. Our formulation naturally handles challenging distributional shift cases (\textit{Concept Drift} and \textit{Target Shift}). We show on several experiments with both synthetic and real world data that our formulation is robust to these cases.

\section{Proposed Approach}

\subsection{Introduction}

\subsubsection{Notations}
We denote a random variable of features $X$ (a realization $x$) and a random variable labels $Y$ (a realization $y$) which respectively take their values in a feature space $\mathcal X$ and a label space $\mathcal Y$. On such spaces, we introduce two probability distributions: $\mathbb P_s$ (source distribution) and $\mathbb P_t$ (target distribution) on $\mathcal X \times \mathcal Y$. We introduce the source domain $\mathcal D_s = (\mathcal X \times \mathcal Y, \mathbb P_s) $ and the target domain $\mathcal D_t = (\mathcal X \times \mathcal Y, \mathbb P_t)$. The expectation over the source domain is noted $\mathbb E^s = \mathbb E_{\mathbb P_s}$ and the target domain $\mathbb E^t = \mathbb E_{\mathbb P_t}$. We call a \textit{Hidden Covariate Representation}, a function $\varphi: \mathcal X \to \mathcal H$ such that $\forall h \in \mathcal H,\mathbb P_s(Y|H=h) = \mathbb P_t(Y|H=h)$. For the purpose of notation, we identify $H$ with $\varphi$ as hidden covariate representation. For a given set $\mathcal A$, the set $\mathcal F_{\mathcal A}$ denotes the set of measurable and bounded functions from $\mathcal A$ to $\mathbb{R}^+$.

\subsubsection{Distribution matching with Maximum Mean Discrepancy}
Invariant Representation methods for learning cross-domain representations rely on the quantification on how data distributions from the source and the target domains differ. Formally, for two given distributions $\mathbb P$ and $\mathbb Q$, such methods introduce a proxy which quantifies how close $\mathbb P$ and $\mathbb Q$ are. In the present work, we suggest to use the Maximum Mean Discrepancy measure \cite{gretton2007kernel, gretton2012kernel} denoted $\mathrm{MMD}(\mathbb P, \mathbb Q)$. Such measure is based on the following property: 
\begin{equation}
    \mathbb P = \mathbb Q \Longleftrightarrow \forall f \in \mathcal F_{\mathcal X}, \mathbb E_{\mathbb P}[f(X)]=  \mathbb E_{\mathbb Q}[f(X)]
\end{equation}
where $\mathcal F_{\mathcal X}$ is the set of measurable and bounded functions of $\mathcal X$. We can derive a proxy of this property called Maximum Mean Discrepancy: 
\begin{equation}
    \mathrm{MMD}_{\mathcal F_{\mathcal X}}(\mathbb P, \mathbb Q) =  \sup_{f \in \mathcal F_{\mathcal X}} \mathbb E_{\mathbb P}[f(X)] - \mathbb  E_{\mathbb Q}[f(X)]
    \label{MMD}
\end{equation}

\subsubsection{Main contributions}
\begin{figure}
    \centering
    \includegraphics[scale=0.25]{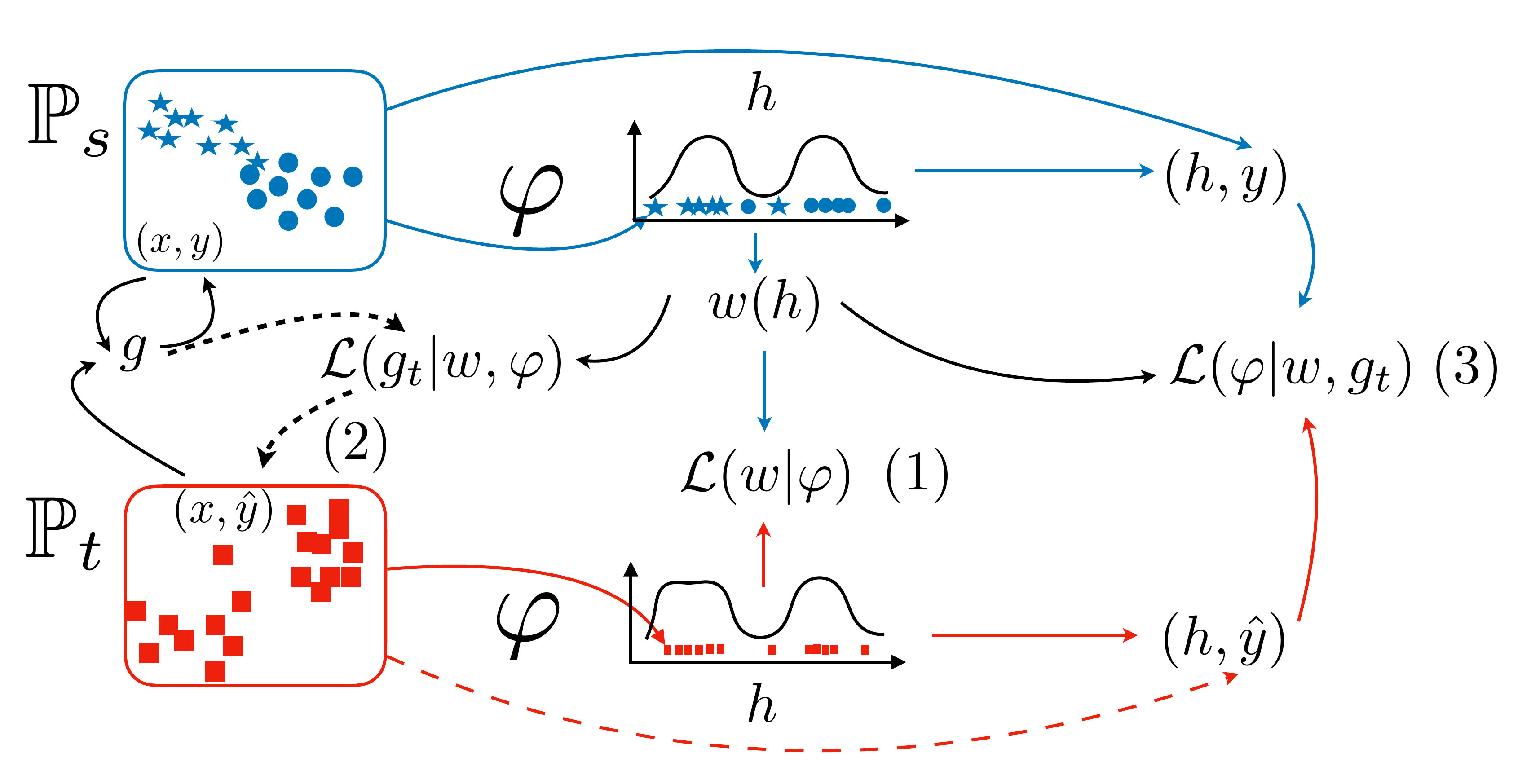}
    \caption{This figure illustrates how our approach works by exhibiting the three learning steps. $(1)$ consists in labeling samples in the target domain with a \textit{Covariate Shift} adaptation in the representation space. $(2)$ consists in evaluating density ratio of representations drawn in the target domain with respect to representations drawn in the source domain. $(3)$ consists in learning $\varphi$ such that $Y|\varphi(X)$ is invariant assuming labeled samples are available in the target domain and density ratio in $(2)$ is exact.}
    \label{fig:global_schema}
\end{figure}
In this section, we derive an optimization procedure from the \textit{Hidden Covariate Shift} assumption. More specifically, we show that a given representation $H = \varphi(X)$ is a \text{Hidden Covariate Representation} if $(\varphi(X), Y)$ in the target domain and a normalized joint distribution $(\varphi(X), Y)$ in the source domain are equal. The normalization consists in a re-weighting by a factor $w(\varphi(X)) =  \mathbb P_t(\varphi(X)) /  \mathbb P_s(\varphi(X))$. This implies the introduction of three different losses which we detail further:
\begin{enumerate}
    \item For a given representation $\varphi$, we need to estimate the density ratio $w(\varphi(X)) =   \mathbb P_t(\varphi(X)) /  \mathbb P_t(\varphi(X))$ (step $(1)$ in Figure \ref{fig:global_schema}). The loss proxy for learning $w$ for a $\varphi$ is denoted $\mathcal L(w |\varphi)$ and called \textit{Hidden Weight Loss}.
    \item Since we address the context of Unsupervised Domain Adaptation, labeled samples in the target domain are not available. We estimate them through a target labeler $g_t$ (step $(2)$ in Figure \ref{fig:global_schema}). We show that learning $g_t$ is equivalent to a Covariate Shift Adaptation in the representation space. The loss proxy for learning $g_t$ for given $\varphi$ and $w$ is noted $\mathcal L(g_t | \varphi, w)$  called \textit{Hidden Covariate Loss}.
    \item Assuming $w$ and labeled samples are available in target domain, it is possible to learn $\varphi$ such that  $\varphi$ is an hidden covariate representation by distribution matching (step $(3)$ in Figure \ref{fig:global_schema}). The loss proxy for learning $\varphi$ for given $w$ and $g_t$ is noted $\mathcal L(\varphi |w , g_t)$  called \textit{Reweighted Distribution Matching Loss}.
\end{enumerate}

\subsection{Formulation}
\subsubsection{From hidden covariate shift assumption to a distribution matching problem}
\label{HCS_to_DM}

Let denote $\varphi$ an Hidden Covariate Representation. Therefore, by definition, for all $x \in \mathcal X$, 
\begin{equation}
    \mathbb P_s(Y |H=\varphi(x) ) = \mathbb P_t(Y|H=\varphi (x)) 
    \label{distribution_match_y}
\end{equation}
The fact that equation \ref{distribution_match_y} holds for all $x \in \mathcal X$ is equivalent to: 
\begin{equation}
\label{depends_x}
    \forall h \in \varphi(\mathcal X), \mathbb E^s_{Y | \varphi(X)=h}[f(y)] =\mathbb E^t_{Y | \varphi(X)=h}[f(y)]
\end{equation}
which holds for all $f \in \mathcal F_{\mathcal Y}$. Since equation \ref{depends_x} holds for all $ h \in \varphi(\mathcal X)$, we can add a dependency with $h$ in $f$\footnote{The choice of a relevant $f$ may differ with a given value of $h$. Considering the case where it exists two $(h_1,h_2) \in \mathcal H^2$ such that $\mathbb P_s(Y|H=h_1) \neq \mathbb P_t(Y|H=h_1)$ and $\mathbb P_s(Y|H=h_2) \neq \mathbb P_t(Y|H=h_2)$. The function $f_1$ which exhibits $\mathbb E^s_{Y|H=h_1}[f_1(y)]\neq \mathbb E^t_{Y|H=h_1}[f_1(y)]$ may be different of the function $f_2$ which exhibits $\mathbb E^s_{Y|H=h_2}[f_2(y)]\neq \mathbb E^t_{Y|H=h_2}[f_2(y)]$}. Thus, equation \ref{depends_x} is equivalent to: 
\begin{equation}
    \label{not_depends}
    \forall f \in \mathcal F,  \mathbb E^s_{Y | H=h}[f(h, y)] =\mathbb E^t_{Y | H=h}[f(h,y)]
\end{equation} 
where $\mathcal F$ is the set of functions $f: \mathcal H \times \mathcal Y \to \mathbb{R}$ such that $\forall h \in \mathcal H, f(h, \cdot)$ is measurable bounded and $\forall y \in \mathcal Y, f(\cdot, y)$ is measurable. Then, it is equivalent to:
\begin{equation}
    \forall f \in \mathcal F, \mathbb E^t_H \mathbb E^s_{Y|H} [f(h,y)] = \underbrace{\mathbb E^t_H \mathbb E^t_{Y |H}[f(h,y)]}_{=\mathbb E^t_{(H,Y)}[f(h,y)]}
\end{equation}
which consists in taking the expectation over the variable $\mathcal \varphi(X)$ for $X \sim \mathbb P_t$. Noting that $\mathbb E^s w = \mathbb E^t $ where $w = \mathbb P_t/  \mathbb P_s$, we obtain:
\begin{equation}
    \forall f \in \mathcal F, \mathbb E^s_{(H,Y)}[w(h) f(h,y)] = \mathbb E^t_{(H,Y)} [f(h,y)]
\end{equation}
The application of the transfer theorem with $\varphi(X) = H$ leads to: 
\begin{equation}
     \mathbb E^s_{(X,Y)}[w(\varphi(x))f(\varphi(x), y)] = \mathbb E^t_{(X,Y)}[f(\varphi(x), y)]
     \label{CHS_to_DM}
\end{equation}
where the equality holds for all $f \in \mathcal F$. To summarize, the \textit{Hidden Covariate Shift} assumption is equivalent to the equality between two families of distributions $(\mathbb P_s(Y|\varphi(X)))_{x \in \mathcal X} $ and $(\mathbb P_t(Y|\varphi(X)))_{x \in \mathcal X}$. We have shown this equality between two distribution families is equivalent to a null discrepancy between the join distribution in the target domain and a re-weighted version in the source domain. To compute this discrepancy, we must evaluate the density ratio $w$. Such a task can be challenging in high dimension and we may want to keep the representation dimension of $\varphi$ reasonably low. Additionally, we need to estimate the label in the target domain.
\label{HCL}
\subsubsection{Losses}

\paragraph{Hidden Covariate Loss}

For a given \textit{Hidden Covariate Representation} $\varphi$, $\mathbb P_s(\varphi(X), Y)$ and $\mathbb P_t(\varphi(X), Y)$ verify the covariate shift assumption introduced in \cite{sugiyama2007covariate}. Authors have shown that Domain Adaptation problem can be solved by instance re-weighting in the loss function of label estimation. In our context, the re-weighting is done in the representation space $\varphi(X)$. Thus, the label estimation $\hat g_t$ in the target domain is obtained by minimizing the following loss: 
\begin{equation}
    \mathcal L(g_t|w, \varphi) = \mathbb E^s_{X,Y} [ w(\varphi(x)) \ell (y, g_t(X)) ]
\end{equation}
where $w(h) = \mathbb P_t(h) / \mathbb P_s(h) $ and $\ell(y,y') = - y \log y '$.

\paragraph{Hidden Weight Loss}
For a given \textit{Hidden Covariate Representation} 
$\varphi$, we have shown in \ref{HCL} that Domain Adaptation in $\varphi(\mathcal X)$ relies on the estimation of $w(h) = \mathbb P_t(h) / \mathbb P_s(h) $. \cite{gretton2007kernel} show that such weights $w$ verifies $\forall f \in \mathcal F_{\mathcal H}, \mathbb E^s_H[w(h) f(h)] = \mathbb E^t_H[f(h)]$ thus an estimation $\hat w$ is obtained by minimizing the following loss:
\begin{equation}
    \mathcal L(w |\varphi) = \sup_{f \in \mathcal F_{\mathcal H}}  \mathbb E^s_H[w(h) f(h)] - \mathbb E^t_H[f(h)]
\end{equation}

\paragraph{Reweighted Distribution Matching Loss}
Assuming that an estimation of both $w$ and $g_t$ are available, from \ref{HCS_to_DM} an estimation $\hat \varphi$ is obtained by minimizing the following loss:
\begin{align}
     \mathcal L(\varphi |w, g_t)\notag = \sup_{f \in \mathcal F} & \{ \mathbb E^s_{(X,Y)}[w(\varphi(x))f(\varphi(x), y)]\\
     &- \mathbb E^t_{X}[f(\varphi(x), \hat y)]\}
\end{align}
where $\hat y(x) = \arg \max_{y \in \mathcal Y} g_t(x)(y)$.
 
\subsection{Learning}
\subsubsection{Unbiased estimation of the losses with RKHS}
Since the supremum on $\mathcal F_{\mathcal X}$ is highly intractable, the work of \cite{gretton2007kernel} suggests to use Reproducing Kernel Hilbert Space (RKHS). For a given unit ball $\mathcal F_k$ of RKHS associated with a kernel $k$, the supremum has a close form:
\begin{equation}
    \arg \max_{f \in \mathcal F_k} = \hat f: x \mapsto \mathbb E_{\mathbb P} [k(x, Y)] -  \mathbb E_{\mathbb Q} [k(x, Y)]
\end{equation}
Furthermore, authors have derived an unbiased empirical estimate of $\mathrm{MMD}_{\mathcal F_k}( \mathbb P, \mathbb Q)$ given $n$ samples $(x_i)_{i=1}^n$ drawn from $\mathbb P$ and $m$ samples $(x_j')_{j=1}^m$ drawn from $\mathbb Q$:
\begin{align}
    \widehat{\mathrm{MMD}_{\mathcal F_k}}( \mathbb P, \mathbb Q)^2 = & \notag \frac{1}{n(n-1)} \sum_{i \neq j}^n k(x_i, x_j) \\
    & \notag + \frac{1}{m(m-1)} \sum_{i \neq j}^m k(x_i', x_j') \\
    & - \frac{2}{mn} \sum_{i,j}^{n,m} k(x_i, x_j')
\end{align}
In our specific context, for a given $w$ and $\varphi$ and noting that $\forall (h,y) \in \mathcal (\mathcal H, \mathcal Y), \mathbb P_{w \cdot s} (h,y) = w(h) \mathbb P_s(h,y)$ ($\mathbb P_{w \cdot s}(h) = w(h) \mathbb P_t(h)$) and $\mathbb P_{\hat t} (h,y) =  \mathbb P_t(\hat y_t |h)) \mathbb P_t(h)$, the \textit{Reweighted Distribution Matching Loss} and the \textit{Hidden Weight Loss} can be expressed as follows: 
\begin{align}
    \mathcal L(\varphi | w, g_s) &= \mathrm{MMD}(\mathcal F, \mathbb P_{w\cdot s}, \mathbb P_{\hat t}) \\
    \mathcal L(w |\varphi) &=  \mathrm{MMD}(\mathcal F_{\mathcal H}, \mathbb P_{w\cdot s}, \mathbb P_{t})
\end{align}
Following the unbiased estimation suggested in  \cite{gretton2007kernel}, considering $(x^s_i, y^s_i)_{i=1}^n \sim \mathbb P_s(X,Y)$ and $(x^t_i)_{i=1}^m$ associated with estimated labels $(\hat y^t_j)_{j=1}^m$, we derive the loss estimators with $h = \varphi(x)$: 
\begin{align}
    \widehat{\mathcal L(\varphi |w)}^2 = & \notag \frac{1}{n(n-1)} \sum_{i \neq j}^n w(h^s_i)w(h^s_j)k((h^s_i,y^s_i),(h^s_j,y^s_j)) \\
    & \notag + \frac{1}{m(m-1)} \sum_{i \neq j}^m k((h^t_i,\hat y^t_i),(h^t_j,\hat y^t_j)) \\
    & - \frac{2}{mn} \sum_{i,j}^{n,m} w(h^s_i)k((h^s_i, y^s_i),(h^t_j,\hat y^t_j))
\end{align}
\begin{align}
\widehat{\mathcal L(w | \varphi)}^2 = & \notag \frac{1}{n(n-1)} \sum_{i \neq j}^n w(h^s_i)w(h^s_j)k(h^s_i,h^s_j) \\
    & \notag + \frac{1}{m(m-1)} \sum_{i \neq j}^m k(h^t_i, h^t_j) \\
    & - \frac{2}{mn} \sum_{i,j}^{n,m} w(h^s_i)k(h^s_i, h^t_j)
\end{align}
Details on kernel estimation are given in Appendix \ref{kernel}


\subsubsection{Optimization procedure}
Using the previous results, the learning of a \textit{Hidden Covariate Representation} $\varphi$ consists in solving the following optimization problem :
\begin{align}
 \hat \varphi  = & \arg \min_{\varphi} \widehat{\mathcal L(\varphi| \hat w, \hat g_t)}
     ~~  \\
   \mbox{such that}   & \label{o_w} ~~~~~~~~~~\hat w = \arg \min_w \widehat{\mathcal L(w | \varphi)}\\
    &  \label{o_g} ~~~~~~~~~~\hat g_t  = \arg \min_{g_t} \mathcal L(g_t | w, \varphi)
\end{align} 
Since the model can collapse to a state where $\varphi(X)$ and $Y$ are independent in the source and in the target domains (which ensures that $\mathbb P_s(Y|\varphi(X)) = \mathbb P_t(\hat Y|\varphi(X))$) we add the loss of the supervised task $\mathcal L(g_s | \varphi) = \mathbb E_{(X,Y)}^s[\ell(y,g_s(\varphi(x)))]$ in target domain as a regularization term: 
\begin{equation}
    \label{o_phi} \hat \varphi  =  \arg \min_{\varphi} \widehat{\mathcal L(\varphi| \hat w, \hat g_t)} + \lambda \inf_{g_s}\mathcal L(g_s |\varphi) \mbox{ s.t. } \ref{o_w}, \ref{o_g}
\end{equation}

We suggest a simple optimization procedure detailed in \ref{alg_training}. We use the notation $\theta^f  =f(\cdot, \theta^f)$ to emphasize that $f$ is parametrized by a set of parameters $\theta^f$. Details about the procedure are given in Appendix \ref{op_detail}.

\subsubsection{Model selection with hidden reversed validation}

Most of Domain Adaptation methods depend on a large set of hyper-parameters. A major difficulty of Domain Adaptation is the lack of labeled data in the target domain. \cite{zhong2010cross} suggest to use the \textit{Reverse Validation} to select the best hyper-parameters. This method consists in training the domain adaptation procedure in order to infer labels on the target domain. Then, the same Domain Adaptation method is used in the reversed source/target situation using estimated labels in the target domain. The best model is finally chosen by comparing these new labels with the ground-truth. Assuming $\mathcal D_s = (\mathcal X \times \mathcal Y, \mathbb P_s)$ and $\mathcal D_t = (\mathcal X \times \mathcal Y, \mathbb P_t)$ is a \textit{Covariate Shift} situation, Domain Adaptation only needs to learn instance weight $w(x) = \mathbb P_t(x) / \mathbb P_s(x)$. It is straightforward to show that the second adaptation only consists in learning  $g_s$, by minimizing $\mathbb E^t_{(X,\hat Y)}\left[\frac{1}{w(x)} \ell(g_s(x),\hat y)\right]$
where $\hat Y$ are estimated labels in the target domain and $w$ the density ratio learned during the first adaptation.  In our context, we aim to learn a function $\varphi$ such that $\mathcal D_s = (\varphi(\mathcal X)\times \mathcal Y, \mathbb P_s)$ and $\mathcal D_t = (\varphi(\mathcal X) \times \mathcal Y, \mathbb P_t)$ is a situation of Covariate Shift. We suggest to apply \textit{Hidden Reverse Validation} which consists in learning $(\varphi, w, g_t)$ with our proposed approach and validate the model with reverse validation in $\varphi(\mathcal X)$ by only learning a new $g_s$ by minimizing $
    \hat g_s = \arg \min_{g_s} \mathbb E^t_{(X,\hat Y)}\left[\frac{1}{w(\varphi(x))} \ell(g_s(\varphi(x)),\hat y)\right] $. \textit{Hidden Reverse Validation} has the major advantage to reduce dramatically computation time since the second adaptation is only an instance re-weighted supervised problem.

\section{Experiments}
\subsection{Baselines}
The literature on Domain Adaptation is large and vastly different methods can be employed (learning invariant features with or without label conditioning). Moreover, several discrepancy measures may be used: $f-$divergence measure (e.g. KL-divergence, Mutual Information), integral divergence measure ($\mathrm{MMD}$) or Optimal Transport based measure (Wasserstein distance\footnote{which is also an integral divergence measure}). For the sake of simplicity and to ensure a fair comparison with methods of the literature, we focus on differences between invariance assumptions. Moreover, the same discrepancy measure is used (i.e. $\mathrm{MMD}$ with the same kernels). Furthermore, we detail how the discrepancy loss of distribution matching is modified for each baseline:

\paragraph{Based on $\varphi(X)$ invariance}
It learns to match $\mathbb P_s(\varphi(X))$ and $\mathbb P_t(\varphi(X))$, thus it assumes $w(\varphi(X))=1$). The distribution matching loss is $\mathcal L(\varphi) = \sup_{f \in \mathcal F_{\mathcal H}} \mathbb E_X^s[f(\varphi(X))] -  \mathbb E_X^t[f(\varphi(X))]$. This invariance covers \cite{ganin2014unsupervised, ganin2016domain, shen2018wasserstein, baktashmotlagh2013unsupervised}

\paragraph{Based on $\varphi(X) |Y$ invariance}
It learns to match $\mathbb P_s(\varphi(X)|Y) = \mathbb P_t(\varphi(X)|\hat Y)$. Thus, $w$ depends only of the labels $w(Y) = \mathbb  P_t(\hat Y) / \mathbb P_s(Y)$. The distribution matching loss is $\mathcal L(\varphi) = \sup_{f \in \mathcal F_{\mathcal H}} \mathbb E_{X,Y}^s[w(Y)f(\varphi(X), Y)] -  \mathbb E_X^t[f(\varphi(X), Y)]$. This invariance covers namely \cite{yan2017mind, manders2018simple, chen2018re, zhang2013domain}.

We will flag our method as based on $Y|\varphi(X)$ invariance.

\subsection{A toy experiment on synthetic data}
\begin{figure}[h!]
\begin{center}
    \includegraphics[scale=0.25]{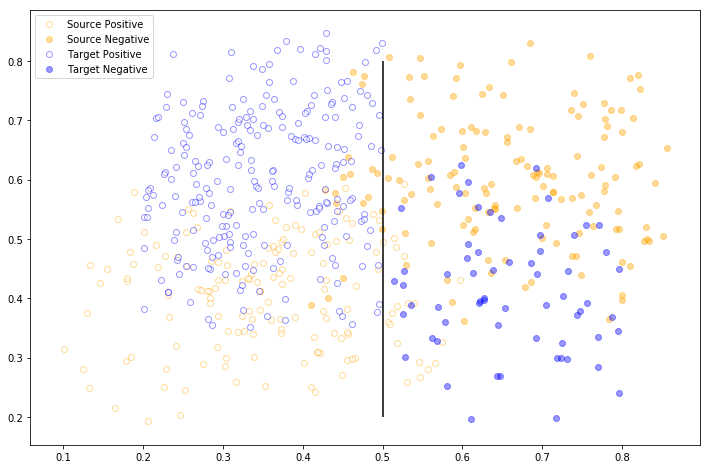}
 
\end{center}

    \begin{center}
        \includegraphics[scale=0.125]{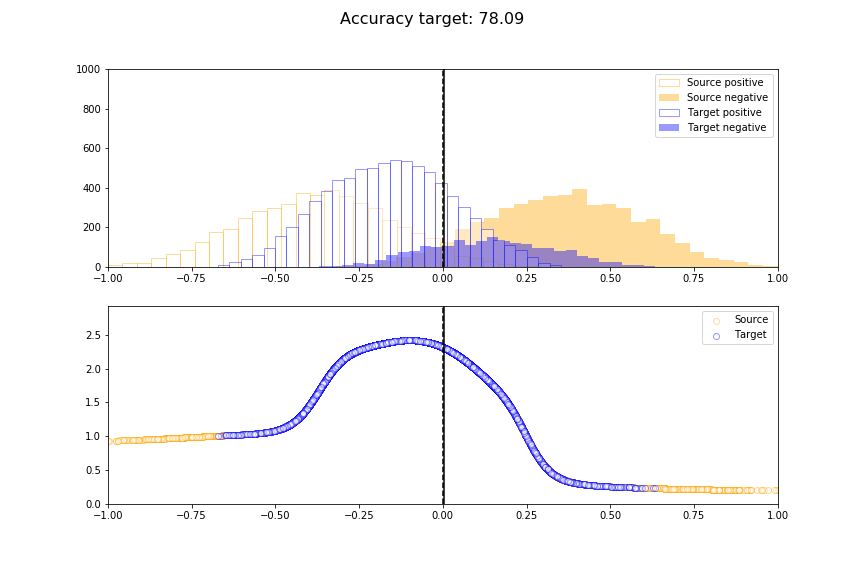}
        \includegraphics[scale=0.125]{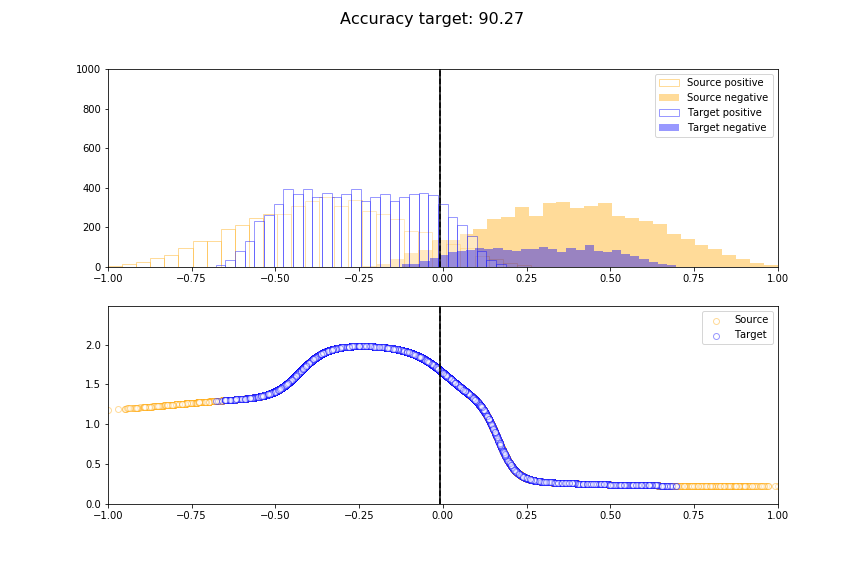}
    \end{center}
    \begin{center}
        \includegraphics[scale=0.125]{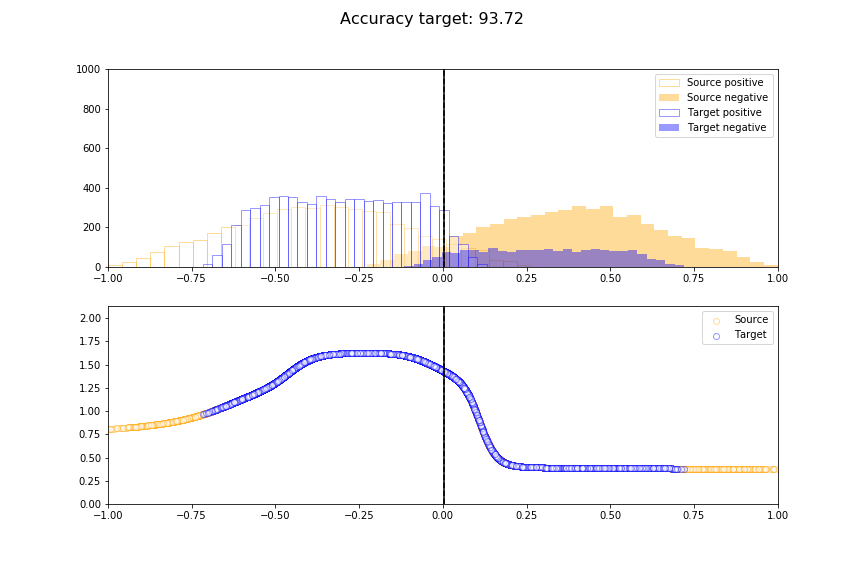}
        \includegraphics[scale=0.125]{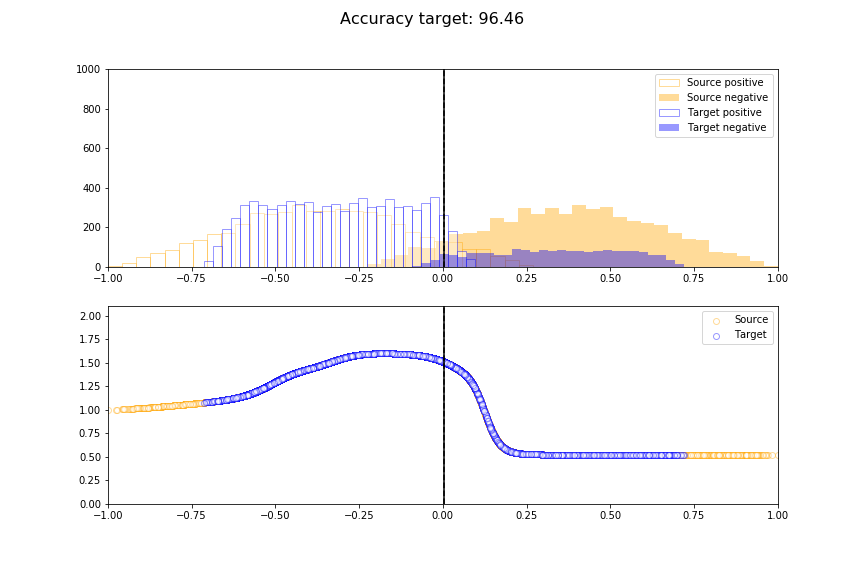}
    \end{center}
    
    \caption{Top: Sample of synthetic data of the toy experiment. The black vertical line is the optimal linear separation. Bottom: four snapshot of learning with our suggested approach reporting $\varphi(X)$ (Top) distribution and $w(\varphi(X))$ estimation (Bottom). The black vertical line is the learned separation in the hidden space.}
    \label{fig:synthetic_data}
\end{figure}

We generate synthetic two dimensional data of a binary classification problem which verifies a challenging case of distributional shift (equations \ref{y}, \ref{xs}, \ref{xt} bellow and Figure \ref{fig:synthetic_data} (Top)). In that context, we observe the situation of \textit{Target Shift} since $\mathbb P_s(Y) \neq \mathbb P_t(Y)$. Besides, it suffers from \textit{Concept Drift} with an inverted $Y$ dependency with $X_2$ between source and target domains. Furthermore $\mathbb P_s(Y|X_1) = \mathbb P_t(Y|X_1)$ with $\mathbb P_s(X_1) \neq \mathbb P_t(X_1)$ but the distribution of $X_1$ is normal in the source domain and uniform in the target domain thus the data verifies the assumption of \textit{Hidden Covariate Shift}. The model we considered is a linear model where features $X$ which lie in two dimensional space are projected to $\varphi(X)$ which lie in a one dimensional space. Then a sigmoid layer is applied on $\varphi(X)$ to determine the label. The estimation of $w(\varphi(X))$ is done with a two layers neural network with $\tanh$ non-linearity and 10 dimensions of hidden states. 

We report in Figure \ref{fig:synthetic_data} the distribution of $\varphi(X)$ and the estimation of $w(\varphi(X))$ for four snapshots equally separated during learning. We observe that after the pretraining step (top left figure), representations from the target domain are not well separated and seem to have a gaussian component. The main reason is the model uses $X_2$ to infer $Y$. During learning (from top left, to top right, to bottom left and to bottom right), we observe that the distribution of  $\varphi(X)$ from the target domain switches from a gaussian distribution to a uniform distribution (which is the distribution of $X_1^t$): the model has learned to forget the feature $X_2$ to infer $Y$ on the target domain. Furthermore, we can observe the estimation of $w$ which integrates both label ratio (to the left of the linear separation, density ratio is much higher than to the right) and the fact that representation from the target domain has a higher density close to the separator than representation from the source domain (since representation in the target domain are harder to separate).
 
\begin{align}
\label{y}
 Y^s &\sim \mathcal B(0.5) ,  ~~ Y^t \sim \mathcal B(0.8) \\
 \label{xs}
 X^s | Y^s &\sim 
 \left ( \begin{array}{ll} X_1^s=  Y^s \mathcal N_t(0.35, 0.125) +  \\
 (1-Y^s) \mathcal N_t(0.65, 0.125) \\
 \hline 
  X_2^s = Y^s \mathcal N_t(0.4, 0.125)  + \\ (1-Y^s) \mathcal N_t(0.6, 0.125)
 \end{array} \right) \\
 \label{xt}
 X^t | Y^t &\sim 
 \left (  \begin{array}{ll}X_1^t =  Y^t \mathcal U(0,0.5) + \\
 (1-Y^t) \mathcal U(0.5,1.0) \\
 \hline 
  X_2^t = Y^t \mathcal N_t(0.6, 0.125) +  \\ 
  (1-Y^t) \mathcal N_t(0.4, 0.125)
 \end{array} \right)
 \end{align}
 
\subsection{Amazon Reviews dataset}
\begin{table}[t]

\vskip 0.15in
\begin{center}
\begin{small}
\begin{sc}
\begin{tabular}{l|c|c|cc}
\toprule
Task & No DA &$\varphi(X)$ & $\varphi(X) | Y$ & $Y|\varphi(X)$ \\
\midrule
E$\to$D & 72.1 & 76.1 &  74.8 & \textbf{75.4} / \textbf{76.1} \\
E$\to$B & 71.8 & 74.4 &  71.8 / 73.3 & \textbf{73.2} / \textbf{74.2}\\
E$\to$K & 83.8 & 86.1 / 87.3 &  \textbf{85.8} / \textbf{86.8} & \textbf{85.4}\\
D$\to$E & 73.6 & 82.2 &  79.8 / \textbf{81.0} & \textbf{80.2} / \textbf{81.0}\\
D$\to$K & 77.3 & 83.0 & \textbf{83.0} & 82.2 / \textbf{83.4}\\
D$\to$B & 78.6 & 78.8 / 79.4 &  77.0 & \textbf{77.9} / \textbf{78.6}\\
B$\to$D & 78.7 & 80.3 &  \textbf{79.7} / \textbf{80.0} & 78.4 / 78.6 \\
B$\to$E & 72.9 & 80.4 / 80.6 &  \textbf{79.2} / \textbf{79.6} & \textbf{78.5} / \textbf{79.0}\\
B$\to$K & 76.1 & 83.7 &  81.8 & \textbf{82.8}\\
K$\to$E &  83.2 & 84.4 &  82.0 / 83.1 & \textbf{83.6} \\
K$\to$D & 73.9 & 80.4 / 80.6 &  \textbf{79.2} / \textbf{79.6} & 78.5 / 79.0\\
K$\to$B & 72.4 & 80.5 / 82.7 &  79.8 / \textbf{81.0} & 80.2 / \textbf{81.0} \\
\midrule 
AVG & 76.2 & 80.9 / 81.2 &   79.5 / 80.1 &   \textbf{79.7}  / \textbf{80.2}\\
\bottomrule
\end{tabular}
\end{sc}
\end{small}
\end{center}
\vskip -0.1in
\caption{Accuracy ($\%$) on Amazon Reviews dataset (standard benchmark). Baseline based on $\varphi(X)$ invariance outperforms significantly approaches based $\varphi(X) | Y$ and $Y|\varphi(X)$. This is due to the prior knowledge on the target distribution $\mathbb P_t(Y) = \mathbb P_s(Y)$.}
\label{standard}
\end{table}

In our experiment on real-world data, we used the Amazon Reviews dataset \cite{blitzer2007biographies}. This dataset contains reviews collected on the Amazon website about products (Book (B), DVD (D), Kitchen (K), Electronics (E)) such that reviews with a rating higher than 4 are considered as positive and considered as negative otherwise. Authors introduced a preprocessed version where relevant unigrams and bigrams are selected. It allows to evaluate the model on 12 tasks of Domain Adaptation ($\to$).  

In order to compare with literature preprocessing choices, for each task, we selected the 5000 most frequent features encoded as a bag of words \cite{ganin2014unsupervised}. The model we used during experiments is a one layer neural networks with 50 dimensions and a $\mathrm{tanh}$ non linearity for learning $\varphi$, a two hidden layer neural network with 10 dimensions for hidden representation for learning $w$, $g_s$ and $g_t$ are sigmoid layers. We used a slow optimizer RMSProp with a learning rate of $0.001$ for learning $\theta_\varphi$ and $0.01$ for learning $\theta_w, \theta_{g_s}$ and $\theta_{g_t}$, a large batch of 128 samples and $\theta_w$ and $\theta_{g_s}$ and $\theta_{g_t}$ are updated 5 times at each update of $\theta_\varphi$. Training is stopped after 30 epochs. Those choices are motivated by numerical stability observed during experiments. The hyper-parameter $\lambda$ (equation \ref{o_phi}) was selected by \textit{Hidden Reversed Validation} among $\{0.1, 1.0, 10.0\}$ repeating the experiment for 3 different but fixed seeds for replicability. Thus, performance on each task is selected among 9 runs. For transparency, we report two results separated with '$/$': left is performance of the model selected by \textit{Hidden Reverse Validation} and right is best performance observed among the 9 runs. This allows to observe the variability of methods. We bold best model comparing methods based on $\varphi(X)|Y$ versus $Y | \varphi(X)$ since methods based on $\varphi(X)$ invariance has a clear advantage since the hypothesis $\mathbb P_s(Y) = \mathbb P_t(Y)$ is verified for the standard benchmark and the \textit{Concept Drift} experiment. We report in Table \ref{standard} the accuracy of our approach compared to baselines on the standard Amazon Reviews (standard benchmark). 


\subsubsection{Concept Drift}

We filtered original datasets for obtaining a \textit{Concept Drift} situation. For a couple of datasets $(\mathcal D_1, \mathcal D_2)$, we build the source domain by over-representing positive reviews from $\mathcal D_1$ and negative reviews from $\mathcal D_2$. The target domain is balanced: it has the same number of negative and positive reviews from each domain. More formally, we sample initial datasets such that for the source domain $\mathbb P_s(Y=1|\mathcal D_1) = 0.8$, $\mathbb P_s(Y=1|\mathcal D_2) = 0.2$ while conserving $\mathbb P_s(Y=1) =0.5$ and $\mathbb P_s(\mathcal D_1) = \mathbb P_s(\mathcal D_2)=0.5$. For the target domain, we took a non-overlapping sample with the source dataset such that $\mathbb P_t(Y|\mathcal D_1) = \mathbb P_t(Y|\mathcal D_2) = 0.5$ and $\mathbb P_t(\mathcal D_1) = \mathbb P_t(\mathcal D_2)=0.5$. The \textit{Concept Drift} situation occurs since determining whether the sample $X$ is drawn from $\mathcal D_1$ or $\mathcal D_2$ helps to infer $Y$ in the source domain but do not help in the target domain. We report results in Table \ref{concept} as the accuracy of the conventional test set of $\mathcal D_2$.

\begin{table}[h!]
\vskip 0.15in
\begin{center}
\begin{small}
\begin{sc}
\begin{tabular}{l|c|c|cc}
\toprule
No DA & Task & $\varphi(X)$ & $\varphi(X) | Y$ & $Y|\varphi(X)$ \\
\midrule
(E, D) & 71.1 & 78.1 & \textbf{76.4} &  75.8  \\
(E, B) & 70.3 & 75.3 / 77.3 &  \textbf{74.0} / \textbf{76.1} & 73.0 / 75.7\\
(E, K) & 83.2 & 86.6 &  84.9 / \textbf{85.4} & \textbf{85.1} / 85.2\\
(D, E) & 76.7 & 83.0 &  80.3 / 82.7 & \textbf{82.1}\\
(D, K) & 75.5 & 85.2 &  \textbf{83.9} / \textbf{85.4} & 82.7 / 83.1\\
(D, B) & 75.4 & 78.0&  75.2 / 76.4 & \textbf{76.3} / \textbf{78.0}\\
(B, D) & 75.0 & 79.3 &  76.5 / 77.9 & \textbf{78.0} / \textbf{78.1}\\
(B, E) & 76.7 & 82.3 &  \textbf{81.4} / \textbf{81.8} & 81.0\\
(B, K) & 77.7 & 85.3 &  82.5 / \textbf{85.2} & \textbf{83.4} / 84.7\\
(K, E) & 81.6 & 82.1 / 83.1  &  80.3 / 82.1 & \textbf{82.4} / \textbf{83.2}\\
(K, D) & 73.7& 76.0 / 76.6 &  74.2 / 76.7 & \textbf{75.6} / 76.3\\
(K, B) & 71.7 & 76.9 &  \textbf{74.6} / \textbf{76.4} & 74.0 / 75.7\\
\midrule 
AVG & 75.7 & 80.7 / 80.9 &  78.7 / \textbf{80.2} & \textbf{79.1} / 79.9\\
\bottomrule
\end{tabular}
\end{sc}
\end{small}
\end{center}
\vskip -0.1in
\caption{Accuracy on Amazon Review dataset in the \textit{Concept Drift} situation. Notation (E, B) stands E $=\mathcal D_1$ and B $= \mathcal D_2$. $\varphi(X)$ invariance provides significant improvements. This is due to the prior knowledge of label distribution in the target domain.}
\label{concept}

\end{table}

\subsubsection{Target shift}

\begin{table*}[t]
\vskip 0.15in
\begin{small}
\begin{center}
\begin{sc}
\begin{tabular}{c|c|c|cc|}
\toprule
Task & No DA &$\varphi(X)$ & $\varphi(X) | Y$ & $Y|\varphi(X)$ \\
\midrule
E$\to$D & 71.2   & 71.6 / 73.0 &  72.6 & \textbf{73.8}\\
E$\to$B & 70.9 & 72.2 &  70.6 / 71.7 & \textbf{72.3}\\
E$\to$K & 85.1 & 81.1 / 81.5 &  83.7 / 85.1 & 
\textbf{84.9} / \textbf{85.9} \\
D$\to$E & 74.2 & 79.0 / 79.5 &  \textbf{83.2} & 80.8\\
D$\to$K & 76.4 & 79.0 / 79.4 &  \textbf{81.3} & 78.1 / \textbf{81.6} \\
D$\to$B & 78.7 & 75.9 / 76.9 & 77.6 & 77.6 \\
B$\to$D & 78.2 & 78.1 &  78.3 / 78.6 & \textbf{79.1}\\
B$\to$E & 71.1 & 76.9 / 77.3 &  \textbf{78.8} / \textbf{80.4} & 77.2 \\
B$\to$K & 76.9 & 81.0 &  \textbf{82.6} & \textbf{82.6} \\
K$\to$E & 82.2 & 81.6 &  \textbf{83.4} & 80.4\\
K$\to$D & 73.1 & 72.9 / 74.4 & 71.8 / 72.0 & \textbf{76.1}\\
K$\to$B & 73.2 & 71.7 / 73.1 &  70.2 / 75.1 & \textbf{73.9} \\
\midrule
AVG & 76.0 & 76.8 / 77.3 &  77.8 / \textbf{78.6} & \textbf{78.1} / 78.4 \\
\bottomrule
\end{tabular}
\begin{tabular}{|c|cc}
\toprule $\varphi(X)$ & $\varphi(X) | Y$ & $Y|\varphi(X)$ \\
\midrule
 69.4 &  71.3 & \textbf{72.8}\\
 69.5 &  66.7 / 67.8 & 69.0 / 69.9\\
 74.8 / 75.0 &  82.9 & 84.0 \\
 71.1/74.8 &  77.1 / 79.1 & \textbf{79.3}\\
  73.4 &  \textbf{80.4} & 78.5 / 78.6 \\
 73.1 / 73.3 & 75.9 & 75.5 \\
 72.8 &  76.8 / 77.7 & 77.3 / \textbf{78.2}\\
  69.8 / 72.5 &  73.0 / \textbf{77.4} & \textbf{75.6} / 76.4\\
 75.3 &  \textbf{79.9} / \textbf{80.5} & 75.4 / 79.2\\
 74.3 &  81.4 / 82.3 & \textbf{83.2}\\
  70.9 & 72.7 / 73.4 & \textbf{73.8} / \textbf{74.1}\\
  72.1 &  70.9 / 71.5 & \textbf{73.3} \\
\midrule
  
 72.2 / 72.8 & 75.8 / 76.7 &  \textbf{76.5} / \textbf{77.0}\\
\bottomrule
\end{tabular}
\end{sc}
\end{center}
\end{small}
\caption{Performances on ACL dataset with \textit{Smooth Target Shift} ($\mathbb P_t(Y=1) = 0.7$) and \textit{Hard Target Shift} ($\mathbb P_t(Y=1) = 0.9$). Method based on $\varphi(X)$ invariance fails to learn in that context while our approach is marginally better than the baseline based on $\varphi(X) | Y$ invariance.}
\label{TS}
\vskip -0.1in

\end{table*}

We filtered original dataset for obtaining a \textit{Target Shift} situation. It consists in rejecting randomly samples in order to obtain a target set with a desired amount of Target Shift. We investigate two cases of target shift: \textit{Soft Target Shift} with $\mathbb P_t(Y=1) = 0.7$ and \textit{Hard Target Shift} with $\mathbb P_t(Y=1) = 0.9$. We report in Table \ref{TS} results of the experiment. 

\subsubsection{Analysis}

We note our approach learns since it improves the model without Domain Adaptation of $+3.3\%$ for the standard benchmark (see Table \ref{standard}). In the situation of \textit{Concept Drift}, our approach still improves the model without Domain Adaptation ($+3.4\%$). In these both cases, our approach remains significantly under the baseline based on $\varphi(X)$ invariance (respectively $-1.2\%$ and $-1.9\%$). This drop of performance is explained by the fact that $\varphi(X)$ invariance baseline has a major advantage since it has a prior knowledge on the target distribution by knowing $\mathbb P_t(Y) = \mathbb P_s(Y)$. Nevertheless, in the context of \textit{Target Shift}, baseline based on $\varphi(X)$ invariance fails to learn in such situation: the model degrades badly compared to the model without Domain Adaptation in the context of \textit{Hard Target Shift} ($-3.8\%$). Our approach still allows to learn in such situation by improving performances with respect to the model without Domain Adaptation (\textit{smooth}: $+2.1\%$, \textit{hard}: $+0.5\%$). For the standard benchmark, the \textit{Concept Drift} and \textit{Target Shift} situation, our approach is (globally) marginally better than the baseline based on $\varphi(X) |Y$ invariance (from $+0.2\%$ for the standard benchmark to $+0.7\%$ for \textit{Hard Target Shift}). This may be explained by the fact our approach can adapt itself to $\varphi(X)|Y$ by learning to set $w(\varphi(X))=w(Y)$ making it a more generic formulation.

\section{Related work}

Addressing \textit{Target Shift} is known as a difficult problem. \cite{manders2018simple} and \cite{yan2017mind} has shown it is possible to adapt respectively the adversarial framework of \cite{ganin2014unsupervised}  and MMD based methods \cite{gretton2007kernel} for learning invariant representation by re-weighting at each training step the domain discrepancy measure with an estimated class distribution in the target domain. In the context of Optimal Transport based discrepancy measure \cite{ arjovsky2017wasserstein, shen2018wasserstein}, \cite{chen2018re} suggests to learn adversarially the class ratio incorporating it into the supremum over the dual critic function of the Wasserstein measure. \textit{Conditional Shift} (where $\mathbb P(X|Y)$ may change while keeping $\mathbb P(Y)$ constant) is also a current assumption for extending invariant representation methods in challenging context of distributional shift. They traditionally use a local-scale transformation for learning such $\psi$ \cite{zhang2013domain}. \cite{gong2016domain} suggests a component-wise version of local-scale transformations for avoiding noisy features which can not be well matched. Those methods are naturally extended in their original work to the context of both target shift and conditional shift using class ratio estimation.

Previous methods are essentially based on marginals or representations conditioned to label matching. However, \textit{Joint Domain Adaptation} methods have recently received a lot of attention and seem well-suited for tackling challenging distributional shift cases. \cite{courty2017joint, damodaran2018deepjdot} have shown it is possible to learn a non linear function $f$ minimizing an Optimal Transport cost between the joint distribution of $(X,Y)$ on the source and the estimated joint distribution $(X, f(Y))$. Furthermore, they successfully adapted it to the context of target shift \cite{redko2018optimal}. \cite{long2016deep, long2018conditional} is the first, to our knowledge, to perform MMD on joint distributions. They suggest to join internal states of a neural network and to perform MMD in the tensor product of reproducing kernel Hilbert spaces associated to each layer. They claim their formulation handles harder distributional shift cases than traditional invariant representation methods since it weights each layer of the network with respect to others in the kernel mean embedding space. Our work mainly differs from the work \cite{long2016deep, long2018conditional} by proposing a constrained re-weighting structure in the kernel mean embedding space. This structural constraint is derived from the hypothesis of \textit{Hidden Covariate Shift} assumption. 

\section{Discussion and Future Work}
In the present work, we have explored a method for performing Unsupervised Domain Adaptation based on the \textit{Hidden Covariate Shift} assumption. To our knowledge, proposed approach is novel and differentiates to current approaches by incorporating during learning the implicit assumption of Domain Adaptation: learning $\varphi(X)$ such that $Y|\varphi(X)$ is conserved.  We adapted the reverse validation method for model selection in our specific case suggesting \textit{Hidden Reverse Validation}. Furthermore, our approach has the interest to take best of two traditional approaches, namely \textit{Covariate Shift} and \textit{Domain Invariant Representation}. We have shown the viability of our formulation in context of \textit{Target Shift} or \textit{Concept Drift} on both synthetic and real-world data. We reported performances comparable to state-of-the-art approaches on Amazon Review dataset. Besides, we have observed that \textit{Hidden Reverse Validation} may not always reflect the performance of selected model at test time on the target set. This may result to the fact we need to estimate a density ratio in high dimension which may be highly noisy.

In a future work, we want to exhibit situation where our formulation has a clear advantage with respect to state-of-the-art approaches. We believe that $Y|\varphi(X)$ invariance is a weaker constraint that $\varphi(X) |Y$ invariance. One the one hand, this allows to learn representations with domain specific information although such information is not used during inference. This can be a desiderata in a context of \textit{Multi-Task Learning} or \textit{Transfer Learning}. On the other hand, this weaker constraint leads to introduce an over-parametrized formulation. Although it is not theoretically justified to look for $\varphi(X) |Y$ invariance for learning $X \to Y$ with the intermediate state $\varphi(X)$, this has the advantage to regularize the model with a receivable heuristic. This intuition was confirmed during experiment when some learnings collapse to learn exactly the ratio of labels distribution (i.e. $w(\varphi(X)) = w(Y)$). Therefore, future work will focus on introducing regularization compatible with the hypothesis of \textit{Hidden Covariate Shift}. Finally, we want to extend our formulation to deep neural network where the network learns a sequence of representation $(H_j)_j$ such that at each layer $H_{j+1} | H_j$ is conserved across domains.

\section*{Acknowledgments} 
This work was funded by Sidetrade and ANRT (France).

\bibliography{example_paper}
\bibliographystyle{icml2018}

\newpage.\newpage

\appendix
\section{Optimization procedure details}
\label{op_detail}

\begin{algorithm}[h!]
   \caption{Learning $\varphi$ hidden covariate representation}
   \label{alg_training}
\begin{algorithmic}
   \STATE {\bfseries Input:} source data $(x^s_i, y^s_i)_{i=1}^n$ sampled from $\mathbb P_s$, target data $(x^t_i)_{i=1}^m$ sampled from $\mathbb P_t$, batch size $b$, $n_w$ weight iterations, $n_g$ iterations.
   \STATE \textbf{initialize} $\theta_\varphi$ with supervision on $(x^s_i, y^s_i)_{i=1}^n$ (\textit{representation pre-training})
   \STATE \textbf{initialize} $\theta_w$ minimizing $\widehat{\mathcal L(\theta_w |\theta_\varphi) }$ (\textit{weight pre-training})
   \REPEAT
   \STATE Sample batch of data of size $b$ from the source / target empirical distributions $(x^s_i, y^s_i)_{i \in \mathcal B^s}$ and $(x^t_i)_{i \in \mathcal B^t}$
   \FOR{$i=1$ {\bfseries to} $n_w$}
   \STATE $\theta_w \leftarrow  \theta_w - \alpha_w \nabla_w \widehat{ \mathcal L(\theta_w | \theta_\varphi)}$
   \STATE $\alpha_w \leftarrow \mbox{LearningRateUpdate}(\alpha_w)$
   \ENDFOR
   \FOR{$i=1$ {\bfseries to} $n_{g_t}$}
   \STATE $\theta_{g_s} \leftarrow  \theta_{g_s}  - \alpha_{g_s}  \nabla_{g_s}  \widehat{ \mathcal L(\theta_{g_s}  |\theta_w, \theta_\varphi)}$
   \STATE $\theta_{g_t} \leftarrow  \theta_{g_t}  - \alpha_{g_t}  \nabla_{g_t}  \widehat{ \mathcal L(\theta_{g_t}  |\theta_w, \theta_\varphi)}$
     \STATE $\alpha_{g_s}  \leftarrow \mbox{LearningRateUpdate}(\alpha_{g_s})$
   \STATE $\alpha_{g_t}  \leftarrow \mbox{LearningRateUpdate}(\alpha_{g_t})$
   \ENDFOR
   \STATE $\theta_{\varphi} \leftarrow  \theta_{\varphi}  - \alpha_{\varphi}  \nabla_{\varphi} \left ( \widehat{ \mathcal L(\theta_{\varphi}  |\theta_w, \theta_{g_t})} + \lambda 
   \mathcal L(g_s |\theta_\varphi) \right)$
   \STATE $\alpha_{\varphi}  \leftarrow \mbox{LearningRateUpdate}(\alpha_{\varphi})$
   \UNTIL{convergence}
\end{algorithmic}
\end{algorithm}

\section{Training details: Kernels}
\label{kernel}
We used the following kernels for computing Maximum Mean Discrepancy: 
\begin{itemize}
    \item Liner kernels, 
    \item Gaussian kernels:
    $$ k_\sigma(x,y) = \exp \left( - \frac{||x-y||^2}{2\sigma^2}\right)$$
    with $\sigma \in \{0.01, 0.1, 0.25, 0.5, 1.0, 2.0, 5.0, 10.0, 50.0, 100.0, 200.0 \}$
    \item Quadratic kernels:
    $$ k_\alpha(x,y) = \left ( 1 + \frac{||x-y||^2}{2 \alpha} \right)^{-\alpha}$$
    with $\alpha \in \{0.01, 0.1, 0.25, 0.5, 1.0, 2.0, 5.0, 10.0, 50.0, 100.0, 200.0 \}$
\end{itemize}
During learning, we need to compute kernel in joint space $\mathcal H \times \mathcal Y$. Since $\mathcal Y$ is discrete, we suggest to cast the variable $(h,y)$ into a continuous variable $(y h, (1-y)h)$.
\end{document}